\documentclass[10pt,twocolumn,letterpaper]{article}

\usepackage{cvpr}              

\newcommand{\red}[1]{{\color{red}#1}}

\definecolor{cvprblue}{rgb}{0.21,0.49,0.74}
\usepackage[pagebackref,breaklinks,colorlinks,allcolors=cvprblue]{hyperref}
\usepackage{iac_pkg}
\def\paperID{119} 
\def\confName{CVPR}
\def\confYear{2025}
\usepackage{multirow}
\title{Retinex-guided Histogram Transformer for Mask-free Shadow Removal}

\author{Wei Dong \quad Han Zhou$^\dagger$ \quad Seyed Amirreza Mousavi \quad Jun Chen$^\dagger$ \\
McMaster University  \quad $^\dagger$Corresponding Authors\\
{\tt\small \{dongw22, zhouh115, mousas27, chenjun\}@mcmaster.ca}
}

\begin{document}


{
\twocolumn[{
\renewcommand\twocolumn[1][]{#1}
\maketitle
\vspace{-6mm}
\begin{center}

\setlength{\abovecaptionskip}{1.5mm}
\setlength{\parskip}{0mm} 
\setlength{\baselineskip}{0mm} 
\begin{minipage}[c]{1\textwidth}
    \includegraphics[width = 1\textwidth]{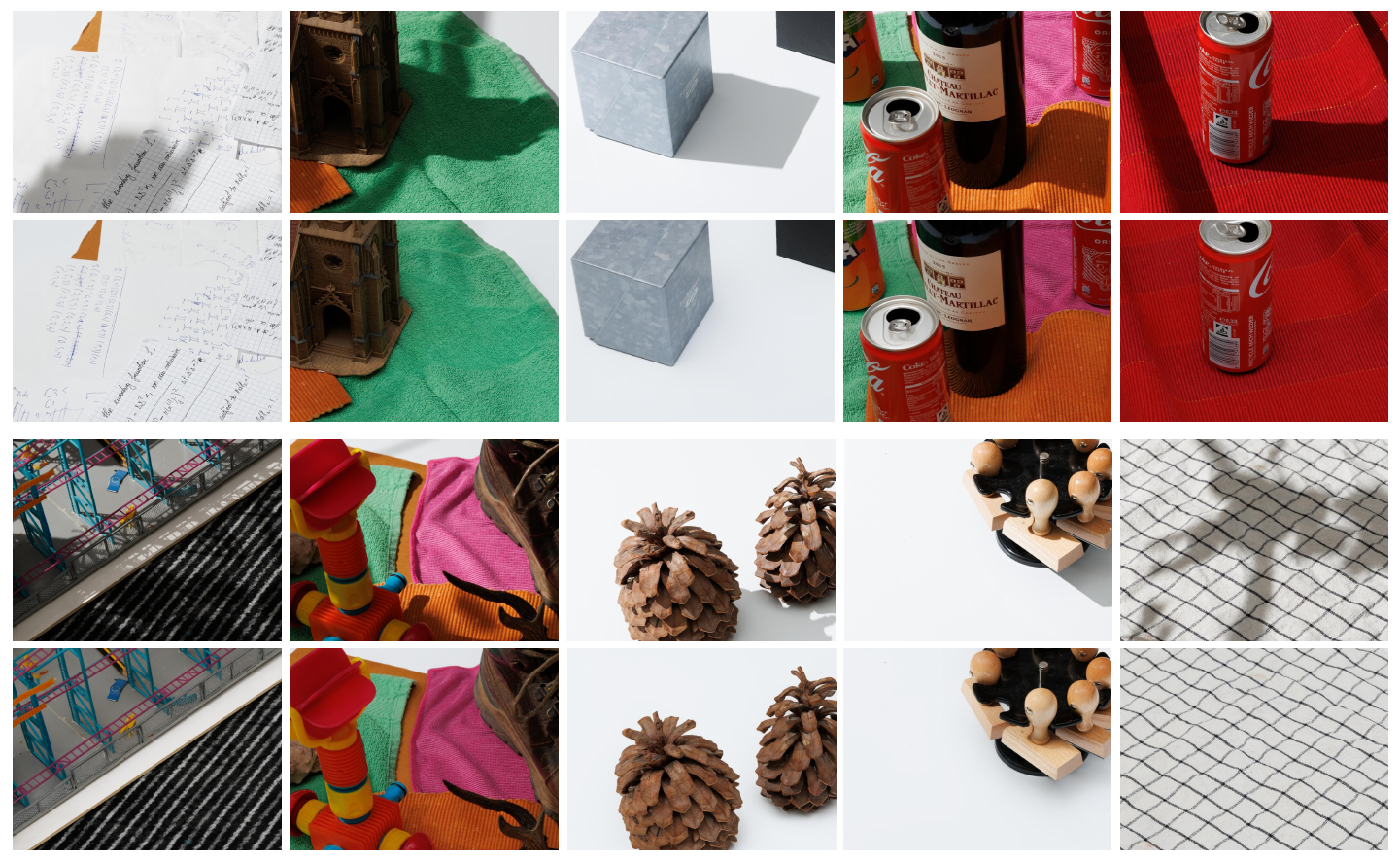}
    \setlength{\parskip}{0mm} 
\end{minipage}
\vspace{-5mm}
\captionof{figure}{Our method demonstrates effective shadow removal from low-quality images and achieves the 7th place in the fidelity track. Notably, it is trained \textbf{exclusively} on the NTIRE 2025 Shadow Removal Challenge training set, and among the top seven solutions, it features \textbf{one of the smallest parameter counts and among the fastest inference speeds}, which underscores its practical applicability under constrained computational resources.}
\vspace{3mm}
\label{fig_first}
\end{center}}]
}

\begin{abstract}
While deep learning methods have achieved notable progress in shadow removal, many existing approaches rely on shadow masks that are difficult to obtain, limiting their generalization to real-world scenes. In this work, we propose \textbf{ReHiT}, an efficient mask-free shadow removal framework based on a hybrid CNN-Transformer architecture guided by Retinex theory. We first introduce a dual-branch pipeline to separately model reflectance and illumination components, and each is restored by our developed Illumination-Guided Hybrid CNN-Transformer (IG-HCT) module. Second, besides the CNN-based blocks that are capable of learning residual dense features and performing multi-scale semantic fusion, multi-scale semantic fusion, we develop the Illumination-Guided Histogram Transformer Block (IGHB) to effectively handle non-uniform illumination and spatially complex shadows. Extensive experiments on several benchmark datasets validate the effectiveness of our approach over existing mask-free methods. Trained solely on the NTIRE 2025 Shadow Removal Challenge dataset, our solution delivers competitive results with one of the smallest parameter sizes and fastest inference speeds among top-ranked entries, highlighting its applicability for real-world applications with limited computational resources. The code is available at \url{https://github.com/dongw22/oath}.
\end{abstract}
    
\section{Introduction}
\label{sec:intro}

Shadows are a common phenomenon in natural scenes, occurring when a light source is partially or fully obstructed by objects, and may result in significant challenges for various high-level computer vision applications (\textit{e.g.}, object tracking, detection and segmentation \cite{tracking, segmentation, recognition}). Consequently, shadow removal has become a fundamental problem in image restoration. 

Before the advent of deep learning, traditional shadow removal methods \cite{HirF8, HirF9, HirF14, HirF18, HirF48} primarily relied on handcrafted discriminative priors to detect and correct shadows based on attributes such as edges, intensities, geometries~\cite{ciconv, lita25cvpr}. Additionally, physics-based illumination models~\cite{retinex} were commonly used to estimate and compensate for lighting differences between shadowed and non-shadowed regions. However, these approaches often struggled in real-world scenarios due to their oversimplified assumptions and the inherent complexity of accurately modeling illumination variations. 

In recent years, learning-based approaches~\cite{HirF6, HirF19, HirF29, HirF30, HirF32, stcgan, refusion, ifblend24eccv, Dong-Shadow} have emerged as a dominant paradigm in shadow removal, leveraging the substantial modeling capacity of deep neural networks. Convolutional neural network (CNN)-based~\cite{ifblend24eccv} and Transformer-based methods~\cite{shadowformer, Dong-Shadow} have demonstrated remarkable success in mapping shadowed images to their shadow-free counterparts through end-to-end learning. These deep learning approaches fall into two broad categories: mask-based \cite{HirF32} and mask-free shadow removal methods~\cite{Dong-Shadow, ifblend24eccv}. Mask-based methods utilize pairs of shadow-affected and shadow-free images alongside explicit shadow masks, either manually annotated or generated by pre-trained models, to guide the learning process. The incorporation of precise shadow masks allows models to focus on learning the complex mapping between shadowed and clean regions, leading to state-of-the-art performance. However, this comes with a cost, these methods face challenges in acquiring accurate shadow masks, especially in complex real-world scenes where manual annotations or automatic predictions may be unreliable. 

On the other hand, current deep learning approaches often fail to fully incorporate the underlying physics of illumination and shadows. Many end-to-end models~\cite{dcshadow}, though effective, struggle with generalization across diverse lighting conditions, leading to artifacts along shadow boundaries. Similarly, physics-based models~\cite{HirF18, HirF48}, despite leveraging certain illumination properties, rely on overly simplistic assumptions such as uniform lighting within shadow regions and basic linear transformations for illumination correction. These limitations motivate the development of more sophisticated shadow removal methods that can effectively handle intricate lighting variations and complex scene geometries. 

To tackle these challenges, we introduce \textbf{ReHiT}, an efficient two-branch mask-free shadow removal network based on illumination-guided hybrid CNN-Transformer architecture. We first extend analyze Retinex theory~\cite{retinex} and develop a Retinex estimator to convert the input into two intermediate representations, each approximating the target reflectance and illumination map~\cite{ecmamba24neurips}.  Second, we present a hybrid CNN-Transformer network, guided by Retinex information, as the core restoration framework of our method. Within each block of this UNet encoder-decoder architecture, we develop the  Illumination-Guided Histogram Transformer Blocks (IG-HTBs) to integrate the illumination guidance and employ the CNN-based  Dilated Residual Dense Block (DRDB) and Semantic-Aligned Scale-Aware Module (SAM) proposed in~\cite{DRDB} multi-scale feature fusion.

Overall, our contribution can be summarized as follows:

\begin{enumerate}
    \item [1.] We introduce \textbf{ReHiT}, an efficient CNN-Transformer hybrid architecture for mask-free shadow removal.
    \item [2.] Guided by Retinex theory, our approach adopts a two-branch restoration pipeline, wherein a hybrid CNN-Transformer network fulfills the restoration process.
    \item [3.] An illumination-guided histogram Transformer is developed to perceive and recover the shadow region in the primary restoration network.
    \item [4.] Experiments across multiple shadow removal benchmarks demonstrate the effectiveness of our method. Moreover, even with less training data and reduced model complexity, our method achieves competitive outcomes in the NTIRE 2025 Shadow Removal Challenge.
\end{enumerate}

\section{Related Work}
\label{sec:RW}

\begin{figure*}[t]
  \centering
  \includegraphics[width=\linewidth]{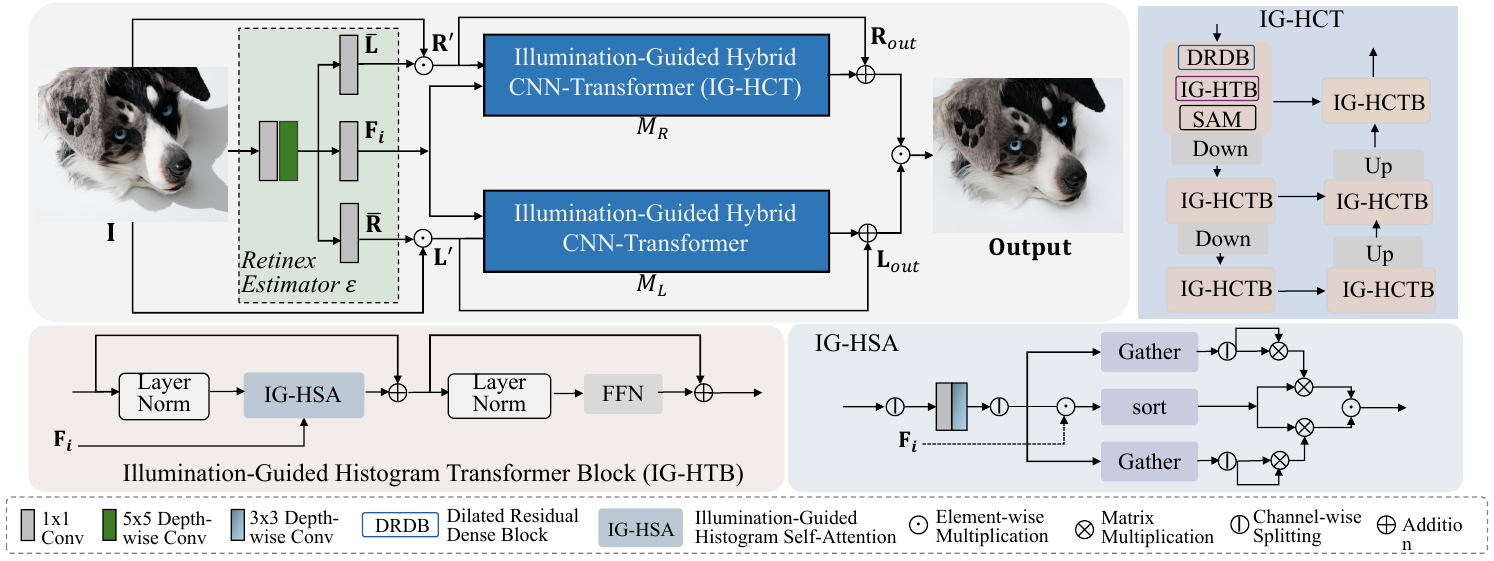}
    \caption{The overall pipeline of our \textbf{ReHiT}. Our solution employs dual-branch Retinex-based pipeline, each branch is dedicated to the restoration of reflectance and illumination map, respectively. The Illumination-Guided Hybrid CNN-Transformers (IG-HCT) module is developed as the primary restoration network. Besides, the Illumination-Guided Histogram Transformer Block (IG-HTB) with the illumination-guided histogram self-attention is combined with the CNN-based Dilated Residual Dense Block (DRDB)~\cite{DRDB} and Semantic-aligned scale-aware module (SAM)~\cite{DRDB} to boost the performance on shadow removal.}
    \label{Fig_frame}
\end{figure*}

\paragraph{Traditional Image Shadow Removal} Early shadow removal methods \cite{HirF8, HirF9, HirF14, HirF18, HirF48} primarily leverage prior knowledge of an image’s physical properties, such as lighting conditions, gradients, regions, and user interactions. Guo et al. \cite{HirF18} reconstruct shadow-free images by modeling illumination relationships between different regions. Finlayson et al. \cite{HirF8, HirF9} employ gradient consistency as a key feature for shadow removal. Gong et al. \cite{HirF14} enhance the robustness of shadow removal algorithms by integrating two user interaction inputs.  Some methods rely on image features and use techniques such as color constancy \cite{DS50}, texture analysis and edge detection \cite{DS2, DS43}. 

\paragraph{Mask-based Image Shadow Removal} In recent years, numerous deep neural models  \cite{HirF6, HirF19, HirF29, HirF30, HirF32, stcgan} are proposed for shadow removal. These methods typically adopt both supervised and unsupervised training strategies and can be categorized into mask-based and mask-free methods. Gryka et al. \cite{DS14} introduce a learning-based approach for automatic shadow removal, utilizing a supervised regression algorithm to handle both umbra and penumbra shadows. ST-CGAN \cite{stcgan} integrates shadow detection and removal using two stacked Conditional Generative Adversarial Networks (CGANs). Wan et al. \cite{HirF39} address the issue of inconsistent static styles between shadowed and shadow-free regions by proposing a style-guided shadow removal network. S2Net \cite{DS3} focuses on semantic guidance and refinement to maintain image integrity. and employs shadow masks to direct shadow removal leveraging semantic-guided blocks to transfer information from non-shadow to shadow regions. He et al. \cite{DS18} develop Mask-ShadowNet, which maintains global illumination consistency through Masked Adaptive Instance Normalization (MAdaIN) and adaptively refines features using aligner modules. Additionally, FusionNet \cite{fusionnet} employs fusion weight maps and a boundary-aware RefineNet to further remove shadow traces. However, these approaches depend heavily on the accuracy of input shadow masks. The complexity and diversity of real-world scenarios make precise shadow mask generation challenging, which may impact the effectiveness of these methods. 

\paragraph{Mask-free Image Shadow Removal} Mask-free methods have shown more flexibility and potential in diverse settings. CANet \cite{DS5} incorporates a Contextual Patch Matching (CPM) module to identify corresponding shadow and non-shadow patches and a Contextual Feature Transfer (CFT) mechanism to transfer contextual information. Vasluianu et al. \cite{ifblend24eccv} introduce Ambient Lighting Normalization (ALN) for enhancing image restoration under complex lighting conditions and propose IFBlend, an advanced image enhancement framework that optimizes Image-Frequency joint entropy, thereby improving visual quality without explicit shadow localization. Le et al. \cite{le2019shadow} leverage an illumination model combined with image decomposition techniques to effectively restore shadow regions. Liu et al. \cite{DS25} introduce a shadow-aware decomposition network, which disentangles illumination and reflectance components, enabling a more precise reconstruction of scene lighting. This is complemented by a bilateral correction network, which refines lighting consistency and restores texture details, ensuring a more natural and perceptually coherent output. ShadowRefiner \cite{Dong-Shadow} is a UNet architecture based on ConvNext~\cite{convnext, zhou2023breaking, dehazedct}, utilizing multi-scale ConvNext blocks as powerful encoders for robust latent feature learning.

\noindent\textbf{Transformer-based Image Restoration.}
Transformer-based networks, which leverage self-attention mechanisms to capture complex relationships between different components, have demonstrated unparalleled efficacy in modeling long-range dependencies~\cite{ntire23dehaze, ntire24dehaze, ntire24shadow, gpp25aaai, ecmamba24neurips, sgllie, ntire2025day, ntire2025lowlight, ntire2025reflection}. Their superior ability to understand contextual relationships has led to state-of-the-art performance in image restoration, surpassing traditional architectures in both accuracy and robustness. SwinIR \cite{DS22}, a widely recognized backbone for image restoration, is constructed using a series of residual Swin Transformer \cite{DS26} blocks, leveraging hierarchical feature representation for enhanced performance. Building upon the Vision Transformer \cite{DS39} framework, DehazeFormer \cite{DS35} has been introduced to address the image dehazing task, demonstrating superior capability in atmospheric degradation removal. Guo et al. \cite{shadowformer} propose Shadowformer to exploit non-shadow regions to help shadow region restoration. More recently, a lightweight transformer architecture \cite{DS4} has been proposed for low-light image enhancement, effectively capturing illumination and reflectance characteristics to improve visual quality under challenging lighting conditions.  In \cite{Dong-Shadow}, a Fast-Fourier attention transformer structure is used in an encoder-decoder architecture to further refine image details and maintain color consistency after shadows are removed.  Sun et al. \cite{HistoFormer} propose a histogram self-attention mechanism to categorize spatial elements into bins and allocate varying attention within and across bins. 
\section{Methodology}
\label{sec:method}

In order to achieve satisfactory shadow removal performance, powerful deep-learning networks are pivotal to extracting important features from shadow-affected images and modeling the mapping from shadow-affected and clean images. The overall framework of our method is shown in Fig.~\ref{Fig_frame}. Our solution employs Retinex theory and an Illumination-Guided Shadow Removal framework with dual pathways \cite{ecmamba24neurips}, each dedicated to the restoration of reflectance and illumination map, respectively (Sec.~\ref{sec_dual_branch}). Retinex theory plays a crucial role in shadow removal by providing a framework for separating reflectance (intrinsic object colors) from illumination variations, which facilitates the refined restoration process of the subsequent Illumination-Guided Hybrid CNN-Transformers (IG-HCT) modules (Sec.~\ref{sec_ighct}). Within each IG-HCT module, the Illumination-Guided Histogram Transformer Block (IG-HTB) with the illumination-guided histogram self-attention is combined with the CNN-based Dilated Residual Dense Block (DRDB) and Semantic-aligned scale-aware module (SAM) to boost the performance (Sec.~\ref{sec_ightb}).


\subsection{Dual-branch Retinex-based Pipeline}
\label{sec_dual_branch}

The Retinex theory can be expressed as $I_{GT} = R_{GT} \odot L_{GT}$, where $I_{GT}$ is an ideal image without shadow, $R_{GT}$ and $L_{GT}$ represent the reflectance image and illumination map, respectively. However, a shadowed image $I_{Sh}$ captured under non-ideal illumination conditions  inevitably suffer from severe noise, color distortion, and constrained contrast.
Therefore as in \cite{ecmamba24neurips}
perturbations ($\hat{R}$ and $\hat{L}$) are introduced to model these shadowed images as:
\begin{gather}
    I_{Sh} = (R_{GT} + \hat{R}) \odot (L_{GT} + \hat{L})\\
    = R_{GT} \odot L_{GT} + R_{GT} \odot \hat{L} + \hat{R} \odot L_{GT} + \hat{R} \odot \hat{L}.
\end{gather}

After introducing $\bar{L}$ and $\bar{R}$ such that  $\bar{L} \odot L_{GT} = 1$  and  $\bar{R} \odot R_{GT} = 1$ and under the assumption that we can approximate $\bar{L}$ and $\bar{R}$ via Retinex estimator, the results can be retrieved using deep learning networks by:
\begin{equation}
\begin{split}
&(\bar{R}, \bar{L}, F_{i}) = \mathcal{E}(I_{Sh}),\\
&R' = I_{Sh} \odot \bar{L}, \text{         } L'=I_{sh}\odot \bar{R},\\
&R_{out} = R' + \mathcal{M}_R(R';F_i),\\ 
&L_{out} = L' + \mathcal{M}_L(L';F_i),\\ 
&I_{out} = R_{out}\odot L_{out},  
\label{eq_retinex}
\end{split}
\end{equation}
where $\mathcal{M}_{R}$ and $\mathcal{M}_{L}$ are networks utilized to predict the minus degradation in $R'$ and $L'$, and $F_i$ serves as a Retinex guidance information derived from the $I_{sh}$.

\subsection{Illumination-Guided Hybrid CNN-Transformers (IG-HCT) Module}
\label{sec_ighct}

Our developed IG-HCT module is an encoder decoder architecture and serves as the $\mathcal{M}_R$ or $\mathcal{M}_L$ in Eq.~\ref{eq_retinex}. This module consists of three down-sampling and up-sampling levels. At each decoder level, the network would produce intermediate results through a convolution layer and a pixelshuffle up-sampling operation, which are also supervised by the ground-truth, serving the purpose of deep supervision to facilitate training. Specifically, each encoder or decoder block, IG-HCTB (top right in Fig.~\ref{Fig_frame}), contains a Dilated Residual Dense Block (DRDB)~\cite{DRDB} for refining the input features, an Illumination Guided Histogram Transformer Block (IG-HTB, introduced in the next subsection) to better capture dynamically distributed shadow-induced degradation,  and a Semantic-aligned multi-scale module (SAM)~\cite{DRDB} for extracting and dynamically fusing multi-scale features at the same semantic level. 

\paragraph{Dilated Residual Dense Block (DRDB)} For each level $i \in \{1, 2, 3, 4, 5, 6\}$ (i.e., three encoder levels and three decoder levels), the input feature $F_i$ first goes through the CNN-based DRDB for refining input features. It incorporates the residual dense block (RDB)~\cite{DRDB46,DRDB15,DRDB14} and dilated convolution layers~\cite{DRDB39} to process the input features and output refined ones. The refined feature representation is then fed to IG-HTB.

\paragraph{Semantic-aligned Multi-scale (SAM) Block} The transformed feature representations generated from IG-HTB is given to the SAM~\cite{DRDB} block to extract multi-scale features within the same semantic level $i$ and allow them to interact and be dynamically fused, significantly improving the model’s ability to handle shadow induced patterns. SAM encompasses two major modules pyramid context extraction and cross-scale dynamic fusion. The pyramid context extraction module first produces pyramid input features given an input feature map through bilinear interpolation and then feeds them into a convolutional branch with five convolution layers to yield pyramid outputs. Given these pyramid features, the cross-scale dynamic fusion module then fuses them together to produce fused multi-scale features to be processed by the next level. 
\setlength\tabcolsep{3pt}
\begin{table*}[ht]
\renewcommand{\arraystretch}{1.3}
\setlength{\abovecaptionskip}{2mm}
\centering

        \scalebox{1}{
        \begin{tabular}{cc|ccc|ccc|ccc}
        \hline
        \multirow{2}{*}{\textbf{Methods}} & \multirow{2}{*}{\textbf{Mask-free}} & \multicolumn{3}{c|}{ISTD~\cite{stcgan}} & \multicolumn{3}{c|}{ISTD+~\cite{le2019shadow}} & \multicolumn{3}{c}{WSRD+~\cite{vasluianu2023wsrd}} \\
\cline{3-11}
        & &PSNR$\uparrow$ &SSIM$\uparrow$ &LPIPS$\downarrow$ &PSNR$\uparrow$ &SSIM$\uparrow$ &LPIPS$\downarrow$   &PSNR$\uparrow$ &SSIM$\uparrow$ &LPIPS$\downarrow$\\ 
        \hline
        DHAN~\cite{dhan} &No &24.86 &0.919 &0.0535 &27.88 &0.917 &0.0529  &22.39 &0.796 &0.1049 \\
        BMNet~\cite{bmnet} &No &29.02 &0.923 &0.0529 &31.85 &0.932 &0.0432 &24.75 &0.816 &0.0948\\
        FusionNet~\cite{fusionnet} &No &25.84 &0.712 &0.3196 &27.61 &0.725 &0.3123 &21.66 &0.752 &0.1227 \\
        SADC~\cite{SADC} &No &29.22 &0.928 &0.0403 &--- &--- &--- &--- &--- &--- \\
        ShadowFormer~\cite{shadowformer} &No &\textbf{30.47} &\textbf{0.928} &\textbf{0.0418} &\textbf{32.78} &\textbf{0.934} &\textbf{0.0385} &\textbf{25.44} &\textbf{0.820} &\textbf{0.0898} \\
        \hline
        DCShadowNet~\cite{dcshadow} &Yes &24.02 &0.677 &0.4423 &25.50 &0.694 &0.4237 &21.62 &0.593 &0.4744 \\
        Refusion~\cite{refusion} &Yes &25.13 &0.871 &0.0571 &26.28 &0.887 &\blue{0.0437} &22.32 &0.738 & 0.0937 \\
        IFBlend*~\cite{ifblend24eccv} &Yes &28.55 &0.906 &0.0558 &30.87 &0.916 & 0.0476 &25.79 &0.809 &0.0905 \\
        ShadowRefiner~\cite{Dong-Shadow} &Yes &\blue{28.75} & \red{0.916} &\red{0.0521} &\blue{31.03} &\red{0.928} &\red{0.0426} &\blue{26.04} &\red{0.827} & \red{0.0854} \\
        
        \hline
        \textbf{Ours (ReHiT)} &Yes &\red{28.81} &\blue{0.914} &\blue{0.0533} &\red{31.16} &\blue{0.925} &0.0442 &\red{26.15} &\blue{0.826} &\blue{0.0860} \\
        \hline
        \end{tabular}
    }
    
\caption{Quantitative comparisons with SOTA methods. Our ReHiT secures comparable performances to ShadowRefiner~\cite{Dong-Shadow} and IFBlend~\cite{ifblend24eccv}, which incorporate large-scale pre-trained ConvNeXt~\cite{convnext} for transfer learning. Compared to mask-based methods, our ReHiT achieves comparable or even better performance (WSRD+ dataset). [Key: \red{Best performance among mask-free models}, \blue{Second-best performance among mask-free models}, \textbf{Best performance among mask-based methods}, *: re-trained with officially released code.]}
\label{table_quant_compar}
\end{table*}

\setlength\tabcolsep{6pt}

\setlength\tabcolsep{4 pt}
\begin{table}[t]
\renewcommand{\arraystretch}{1.3}
    \setlength{\abovecaptionskip}{2mm}
    \centering
    \scalebox{1}{
    \begin{tabular}{c|cc}
    \hline
        
    \textbf{Methods} & Params (M) $\downarrow$  &FLOPs (G) $\downarrow$ \\ 
    \hline
    DCShadowNet~\cite{dcshadow}  & 30.6    &351.0  \\
    ShadowRefiner~\cite{Dong-Shadow} & 293.8   & 274.7  \\
    IFBlend~\cite{ifblend24eccv}   & 271.9 \  &103.8  \\
    \hline
    \textbf{ReHiT (Ours)}  & \red{17.5}   & \red{66.4} \\
    \hline
    \end{tabular}
    }    
\caption{Model complexity comparison between our ReHiT and other mask-free methods. Our ReHiT presents significantly enhanced efficiency and offers more application potentials.}
\label{tab_complexity}
\end{table}
\setlength\tabcolsep{6pt}

\subsection{Illumination-Guided Histogram Transformer Block (IG-HTB)}
\label{sec_ightb}

As the core element of our IG-HCT module, IG-HTB compromises of two essential mechanisms: IG-HSA and FFN. These components are structured to engage with layer normalization and can be expressed as the following.
\begin{equation}
\begin{split}
  F_i &= F_{i-1} + \textit{IG-HSA}(LN(F_{i-1})),\\
  F_i &= F_i  + FFN(LN(F_i )),
\end{split}
\end{equation}
where $LN(\cdot)$ denotes layer normalization and $F_i$ represents the feature at $i$-th level.

To more effectively capture shadow-induced degradation that varies dynamically, we develop an illumination-guided Histogram Self-Attention (IG-HSA) mechanism. This layer incorporates a dynamic-range convolution process, which reorganizes the spatial arrangement of fractional features, along with a histogram self-attention mechanism that integrates both global and local dynamic feature aggregation. Traditional convolution, which primarily focuses on local information, does not naturally complement the self-attention mechanism’s capability to model long-range dependencies. To address this limitation, we use a dynamic-range convolution approach that restructures input features before applying standard convolution operations. Moreover, the illumination information extracted in Sec.~\ref{sec_dual_branch} is integrated to modulate the attention calculation process.

Contrary to most existing vision Transformers~\cite{HistoF11,HistoF75,HistoF78,HistoF86,HistoF96} which leverage fixed range of attention which restricts the self-attention to span adaptively long range to associate desired features, we have noticed that shadow-induced degradation had better be assigned with various extent of attention. We thus propose a histogram self-attention mechanism to categorize spatial elements into bins and allocate varying attention within and across bins. For the sake of parallel computing, we set each bin contains identical number of pixels during implementation.

\begin{figure*}[t]
    \setlength{\abovecaptionskip}{1mm}
    \centering
    \includegraphics[width=1\textwidth]{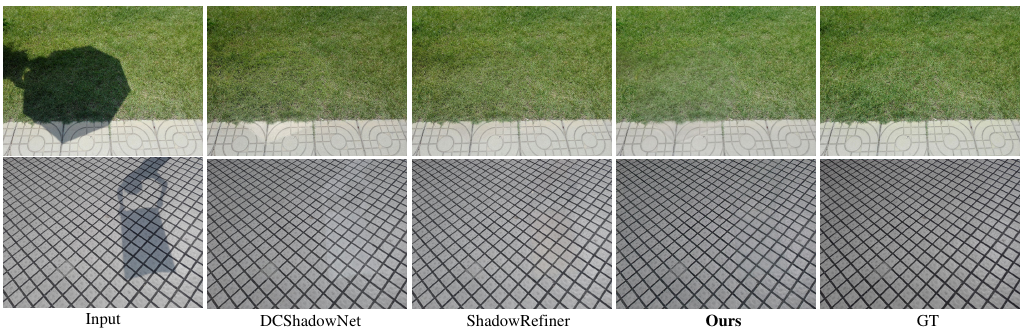}
    \caption{Visual comparisons on the ISTD dataset~\cite{stcgan}. DCShadow preserves textures better but fails to completely eliminate shadows, resulting in visible residuals and boundary artifacts. Similar to ShdowRefiner, our method successfully remove shadows without incorporating artifacts, and produces more uniform and natural shadow removal results, effectively eliminating both soft and hard shadows while preserving underlying textures.}
    \label{ISTD}
\end{figure*}
\begin{figure*}[t]
    \setlength{\abovecaptionskip}{1mm}
    \centering
    \includegraphics[width=1\textwidth]{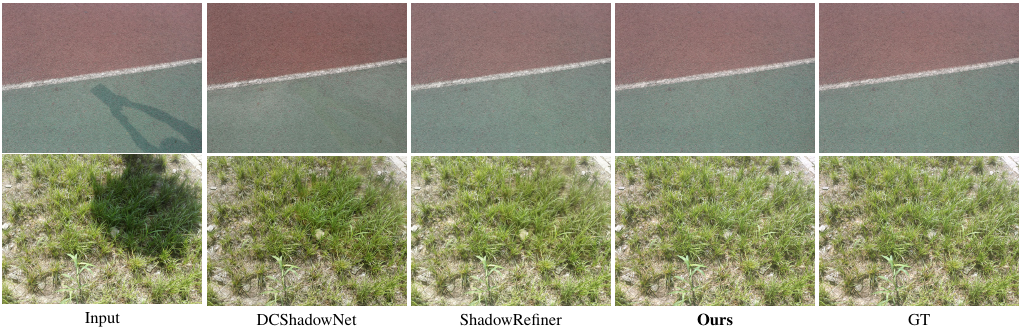}
    \caption{Visual comparisons on the ISTD+ dataset~\cite{le2019shadow}. These results indicate that DCShadow partially remove the shadows but leave noticeable residuals and illumination inconsistencies, especially along shadow boundaries. In contrast, both ShadowRefiner and our method more effectively eliminates shadows with minimal artifacts, producing results that are visually closer to the ground truth. Notably, we only leverage 5$\%$ parameters of ShadowRefiner to achieve comparable performance.}
    \label{ISTD+}
\end{figure*}


\section{Experiments}
\label{sec:exp}

\subsection{Datasets and Implementation Details} 
\paragraph{Datasets} We evaluate our method on three benchmark datasets: (1) The ISTD \cite{stcgan} dataset, (2) The Adjusted ISTD (ISTD+) \cite{le2019shadow} dataset, which reduces illumination inconsistencies between shadow and shadow-free images in ISTD using an image processing algorithm, and (3) The WSRD+ \cite{vasluianu2023wsrd} dataset, where $1000$ training pairs are utilized for training, $100$ validation pairs are used for validation. 

\paragraph{Implementation details} We train our method on a single RTX 3090Ti GPU. Data augmentation techniques include random rotations of $90^\circ$, $180^\circ$, or $270^\circ$, and vertical and horizontal flipping. The crop size and batch size is set to $384\times384$ and $4$, respectively. Optimization is conducted using the Adam optimizer with default hyper-parameters ($\beta_1 = 0.9$, $\beta_2 = 0.999$). The learning rate starts at $1 \times 10^{-4}$ and is gradually reduced to $6.25 \times 10^{-6}$. Besides the L1 and multi-scale SSIM loss~\cite{glare24eccv}, the structure loss~\cite{lita25cvpr} and additional constraints~\cite{ecmamba24neurips} are used for optimization supervision.
 
\begin{figure*}[t]
    \setlength{\abovecaptionskip}{1mm}
    \centering
    \includegraphics[width=1\textwidth]{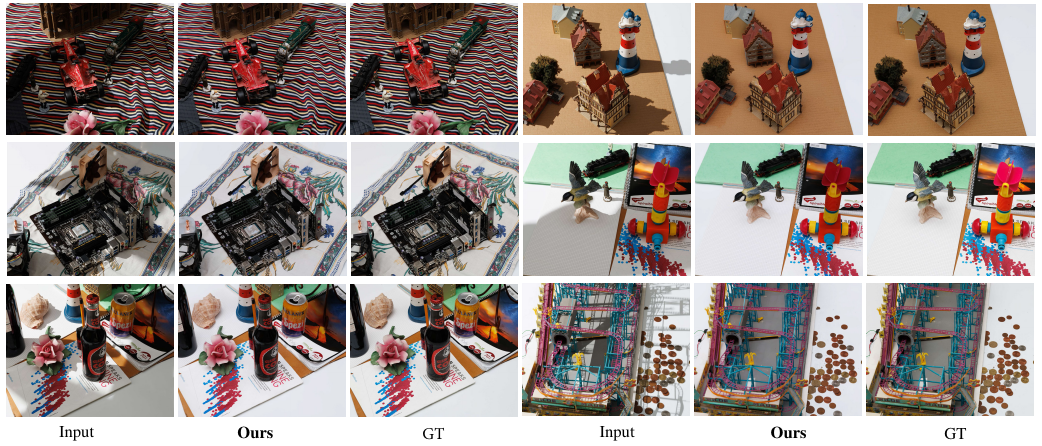}
    \caption{Our method delivers promising performance on the WSRD+ validation set~\cite{vasluianu2023wsrd}. Our method effectively removes both soft and hard shadows, while preserving fine structural and textural details. Notably, regions previously affected by shadows are relit with high fidelity and without introducing noticeable artifacts. The outputs demonstrate consistent illumination and seamless transitions between previously shadowed and non-shadowed areas, highlighting the robustness and generalization ability of our approach in handling real-world shadow removal.}
    \label{WRSD+}
\end{figure*}

\paragraph{Evaluation Metrics}
To fully assess the effectiveness of various shadow removal methods, we employ three quantitative metrics: Peak Signal-to-Noise Ratio (PSNR), Structural Similarity Index (SSIM) \cite{DS42}, and Learned Perceptual Image Patch Similarity (LPIPS) \cite{DS49}, which measure both pixel-level accuracy and perceptual quality. 

\subsection{Comparison with Other Methods} 

We compare our proposed method with several State-of-the-art (SOTA) algorithms. Specifically, several mask-free methods including Refusion~\cite{refusion}, DCShadowNet~\cite{dcshadow}, ShadowRefiner~\cite{Dong-Shadow}, IFBlend~\cite{ifblend24eccv} and several mask-based approaches including ShadowFormer~\cite{shadowformer}, SADC~\cite{SADC} are adopted for comparison. 

\paragraph{Quantitative Results} As documented in Tab.\ref{table_quant_compar}, our approach consistently achieves top-tier performance among mask-free methods across ISTD\cite{stcgan}, ISTD+\cite{le2019shadow}, and WSRD+\cite{vasluianu2023wsrd}. It attains improved PSNR and comparable SSIM, LPIPS values compared to ShadowRefiner~\cite{Dong-Shadow}, the winner solution of NTIRE 2024 Image Shadow Removal Challenge. Although mask-based models typically hold an advantage due to the use of shadow masks, our method achieves comparable accuracy on ISTD and ISTD+ without mask input. On WSRD+, where only estimated masks~\cite{dhan} are available, our method outperforms the best mask-based model (ShadowFormer~\cite{shadowformer}), illustrating its generalization capability in real-world scenarios.

\paragraph{Qualitative Comparisons} Visual comparisons on ISTD dataset, ISTD+ dataset and WSRD+ dataset are reported in Fig.~\ref{ISTD},~\ref{ISTD+}, and ~\ref{WRSD+}, respectively. Our method produces results that closely resemble the ground truth in both color fidelity and detail preservation. For example, in Fig. \ref{ISTD} (row 2), shadows are effectively removed without introducing artifacts or color shifts—issues frequently encountered with competing methods. Similarly, in Fig. \ref{ISTD+} (row 2), our method successfully brightens shadowed regions on the court, seamlessly aligning them with sunlit areas while preserving texture and chromatic consistency.  

As shown in Table~\ref{tab_complexity}, our method requires fewer parameters and FLOPs at inference. Combined with its competitive quantitative performance and the vivid, high-quality results restored by ReHiT in Fig.\ref{ISTD}, \ref{ISTD+}, and \ref{WRSD+}, this demonstrates that our method delivers comparable or even better results with substantially lower computational overhead, indicating its practical value in real-world scenarios.

\begin{table*}[!h]
    \setlength{\abovecaptionskip}{2mm}
    \centering
    \renewcommand{\arraystretch}{1.15}
    \scalebox{1}{
    \begin{tabular}{c|cccc|c}
    \hline
    
    \textbf{Team} & PSNR$\uparrow$ & SSIM$\uparrow$ & LPIPS$\downarrow$ & Params$\downarrow$ & Fianl Rank$\downarrow$  \\ 
    \hline
        X-Shadow     &25.336 &0.839 &0.078 &235 &1  \\
        LUMOS       &25.377 & 0.838 & 0.095 &23 &2  \\
        ACVLab    & 25.895 & 0.842 & 0.124 & 247 &3  \\
        GLHF      & 25.309 & 0.833 & 0.096 & 379 &4   \\
        FusionShadowRemoval  & 24.805 & 0.825 & 0.090 & 373 &5   \\
        LVGroup\_HFUT   & 25.020 & 0.833 & 0.109 & 17 &6   \\
        \textbf{Oath (ours)}   & 24.638 & 0.829 & 0.106 & 17.5 &\textbf{7}   \\
        Alchemist  & 23.958 & 0.817 & 0.088 & 281 &8   \\
        ReLIT     & 23.441 & 0.822 & 0.113 & 175.26 &9  \\
        MIDAS   & 24.439 & 0.819 & 0.130 & 293.83 &10  \\
         KLETech-CEVI   & 23.457 & 0.815 & 0.127 & 10 &11  \\
         PSU Team    & 24.149 & 0.811 & 0.133 & 41 & 12 \\
MRT-ShadowR    & 22.147 & 0.780 & 0.118 & 2.39 & 13 \\
CV\_SVNIT   & 23.866 & 0.772 & 0.216 & 4.7  & 14 \\
X-L    & 21.454 & 0.781 & 0.149 & 5  & 15\\
 ZhouBoda   & 14.857 & 0.555 & 0.407 & 67.07 & 16 \\
  GroupNo9  & 17.043 & 0.550 & 0.660 & N/A & 17 \\  
        \hline
        \end{tabular}
    }
\caption{Final ranking of NTIRE 2025 Shadow Removal Challenge. While maintaining a certain level of performance, \textbf{\textit{our model remains significantly more lightweight}} compared to those achieving top performance. This leads to advantages such as improved deployment efficiency, reduced computational overhead, and enhanced applicability in real-time or resource-constrained environments.}    
\label{lab_rank}
\end{table*}

\subsection{Performance on NTIRE 2025 Shadow Removal Challenge } 
According to the challenge report~\cite{ntire2025shadow} and Tab.~\ref{lab_rank}, our method consistently achieves top-7 performance across all fidelity-related metrics, demonstrating the effectiveness of our method on shadow removal task. Remarkably, our method is trained solely on the NTIRE 2025 Shadow Removal Challenge training set, and among the leading solutions, our model stands out for maintaining one of the smallest parameter sizes and fastest inference speeds among top entries, underscoring its efficiency and practicality in resource-constrained environments. Besides, according to the challenge report~\cite{ntire2025shadow}, almost all solutions with higher performance utilize extra data for training, or includes more training parameters, or presents higher inference latency.

We also report the result of our method for the official test data used in NTIRE 2025 Shadow Removal Challenge as Fig.~\ref{fig_first}, where our method demonstrates consistently favorable shadow removal results across a wide range of scenes.
As illustrated in the first image of Fig.~\ref{fig_first}, the shadow is thoroughly removed. Besides, the color consistency between objects previously occluded by shadows and those directly illuminated highlights the high fidelity of color restoration. Moreover, these results appear under uniformly soft lighting, reflecting ours method’s effectiveness in achieving global illumination consistency.

\subsection{Ablation Study}
\label{sec_ex_abla}

\setlength\tabcolsep{3.2pt}
\begin{table}[t]
\renewcommand{\arraystretch}{1.3}
    \setlength{\abovecaptionskip}{2mm}
    \centering
    \scalebox{1}{
    \begin{tabular}{c|ccc}
    \hline
        
    \textbf{Configurations} &PSNR$\uparrow$  &SSIM$\uparrow$ &LPIPS$\downarrow$ \\ 
    \hline
    \textbf{Full Model} &\red{\textbf{26.15}} &\red{\textbf{0.826}} &\red{\textbf{0.0860}} \\
    w/o dual-branch pipeline &25.86 &0.818 &0.0893 \\
    w/o IG-HTB &25.74 &0.816 &0.0915 \\
    w/o illumination in IG-HTB &25.97 &0.821 &0.0872 \\    
    \hline
    \end{tabular}
    }    
\caption{The ablation results on the WRSD+ dataset demonstrate that each component of our method contributes to the overall effectiveness in shadow removal.}
\label{tab_abla}
\end{table}
\setlength\tabcolsep{6pt}
This section presents a series of ablation studies conducted on the WSRD+ dataset~\cite{vasluianu2023wsrd}.

\paragraph {Importance of Dual-branch Retinex-based Pipeline} To study the contribution of the dual-branch Retinex-based pipeline discussed in Sec.~\ref{sec_dual_branch}, we first remove this pipeline to figure out its contribution for the overall performance of our method provided in Tab.~\ref{table_quant_compar}. Specifically, we directly employ our developed hybrid CNN-Transfomer module to learn the mapping form shadowed images to clean images. The quantitative result is reported in Tab.~\ref{tab_abla}. It demonstrates removing dual-branch Retinex-based pipeline leads to obviously degraded performance, underscoring its importance of for satisfactory shadow removal performance.

\paragraph {Contributions of IG-HTB and the Illumination Guidance} We also implement several further adaptations to illustrate the effectiveness of our developed Illumination-Guided Histogram Transformer Block (IG-HTB) and the illumination guidance in IG-HTB. By separately comparing row 3 and row 4 to our full model in Tab.~\ref{tab_abla}, we can conclude that the design the IG-HTB and the interpolation of illumination guidance help achieve more pleasant performance.

\section{Conclusion}
In this paper, we proposed a lightweight, mask-free shadow removal framework that integrates CNNs and Transformers under the guidance of Retinex theory. By decomposing input images into reflectance and illumination components, our method effectively models illumination inconsistencies caused by shadows. The introduction of the Illumination-Guided Histogram Transformer enhances the network’s ability to perceive and correct complex shadow artifacts. Experimental results across multiple benchmarks show that our method achieves reasonably good performance while maintaining a significantly lower model complexity. In the NTIRE 2025 Shadow Removal Challenge, our approach ranked 7th out of 17 teams, demonstrating its practicality and competitiveness under constrained training resources. These results suggest that our design offers a promising trade-off between performance and efficiency for real-world shadow removal applications.

\clearpage
{
    \small
    \bibliographystyle{ieeenat_fullname}
    \bibliography{main}

\begin{thebibliography}{69}
\providecommand{\natexlab}[1]{#1}
\providecommand{\url}[1]{\texttt{#1}}
\expandafter\ifx\csname urlstyle\endcsname\relax
  \providecommand{\doi}[1]{doi: #1}\else
  \providecommand{\doi}{doi: \begingroup \urlstyle{rm}\Url}\fi

\bibitem[Ancuti et~al.(2023)Ancuti, Ancuti, Vasluianu, Timofte, Zhou, Dong, Liu, Chen, Liu, Li, et~al.]{ntire23dehaze}
Codruta~O Ancuti, Cosmin Ancuti, Florin-Alexandru Vasluianu, Radu Timofte, Han Zhou, Wei Dong, Yangyi Liu, Jun Chen, Huan Liu, Liangyan Li, et~al.
\newblock Ntire 2023 hr nonhomogeneous dehazing challenge report.
\newblock In \emph{Proceedings of the IEEE/CVF Conference on Computer Vision and Pattern Recognition}, pages 1808--1825, 2023.

\bibitem[Ancuti et~al.(2024)Ancuti, Ancuti, Vasluianu, Timofte, Liu, Wang, Zhu, Shi, Lu, Fu, et~al.]{ntire24dehaze}
Codruta~O Ancuti, Cosmin Ancuti, Florin-Alexandru Vasluianu, Radu Timofte, Yidi Liu, Xingbo Wang, Yurui Zhu, Gege Shi, Xin Lu, Xueyang Fu, et~al.
\newblock Ntire 2024 dense and non-homogeneous dehazing challenge report.
\newblock In \emph{Proceedings of the IEEE/CVF Conference on Computer Vision and Pattern Recognition}, pages 6453--6468, 2024.

\bibitem[Arbel and Hel-Or(2010)]{DS2}
Eli Arbel and Hagit Hel-Or.
\newblock Shadow removal using intensity surfaces and texture anchor points.
\newblock \emph{IEEE Transactions on Pattern Analysis and Machine Intelligence}, 33\penalty0 (6):\penalty0 1202--1216, 2010.

\bibitem[Bao et~al.(2022)Bao, Liu, Gang, Yang, and Liao]{DS3}
Qiqi Bao, Yunmeng Liu, Bowen Gang, Wenming Yang, and Qingmin Liao.
\newblock S2net: Shadow mask-based semantic-aware network for single-image shadow remova.
\newblock \emph{IEEE Transactions on Consumer Electronics}, 68\penalty0 (3):\penalty0 209--220, 2022.

\bibitem[Cai et~al.(2023)Cai, Bian, Lin, HaoqianWang, Timofte, and Zhang]{DS4}
Yuanhao Cai, Hao Bian, Jing Lin, HaoqianWang, Radu Timofte, and Yulun Zhang.
\newblock Retinexformer: One-stage retinexbased transformer for low-light image enhancement.
\newblock In \emph{Proceedings of the IEEE/CVF international conference on computer vision}, pages 12504--12513, 2023.

\bibitem[Chen et~al.(2023)Chen, Li, Li, and Pan]{HistoF11}
Xiang Chen, Hao Li, Mingqiang Li, and Jinshan Pan.
\newblock Learning a sparse transformer network for effective image deraining.
\newblock In \emph{Proceedings of the IEEE/CVF conference on computer vision and pattern recognition}, pages 5896--5905, 2023.

\bibitem[Chen et~al.(2021)Chen, Long, Zhang, and Xiao]{DS5}
Zipei Chen, Chengjiang Long, Ling Zhang, and Chunxia Xiao.
\newblock Canet: A context-aware network for shadow removal.
\newblock In \emph{Proceedings of the IEEE/CVF international conference on computer vision}, pages 4743--4752, 2021.

\bibitem[Cun et~al.(2020)Cun, Pun, and Shi]{dhan}
Xiaodong Cun, Chi-Man Pun, and Cheng Shi.
\newblock Towards ghost-free shadow removal via dual hierarchical aggregation network and shadow matting gan.
\newblock In \emph{AAAI}, 2020.

\bibitem[Ding et~al.(2019)Ding, Long, Zhang, and Xiao]{HirF6}
Bin Ding, Chengjiang Long, Ling Zhang, and Chunxia Xiao.
\newblock Argan: Attentive recurrent generative adversarial network for shadow detection and removal.
\newblock In \emph{Proceedings of the IEEE/CVF international conference on computer vision}, pages 10212--10221, 2019.

\bibitem[Dong et~al.(2018)Dong, Zhou, and Xu]{segmentation}
Wei Dong, Han Zhou, and Dong Xu.
\newblock A new sclera segmentation and vessels extraction method for sclera recognition.
\newblock In \emph{2018 10th International Conference on Communication Software and Networks (ICCSN)}, 2018.

\bibitem[Dong et~al.(2024{\natexlab{a}})Dong, Zhou, Tian, Sun, Liu, Zhai, and Chen]{Dong-Shadow}
Wei Dong, Han Zhou, Yuqiong Tian, Jingke Sun, Xiaohong Liu, Guangtao Zhai, and Jun Chen.
\newblock Shadowrefiner: Towards mask-free shadow removal via fast fourier transformer.
\newblock In \emph{Proceedings of the IEEE/CVF Conference on Computer Vision and Pattern Recognition}, pages 6208--6217, 2024{\natexlab{a}}.

\bibitem[Dong et~al.(2024{\natexlab{b}})Dong, Zhou, Wang, Liu, Zhai, and Chen]{dehazedct}
Wei Dong, Han Zhou, Ruiyi Wang, Xiaohong Liu, Guangtao Zhai, and Jun Chen.
\newblock Dehazedct: Towards effective non-homogeneous dehazing via deformable convolutional transformer.
\newblock In \emph{Proceedings of the IEEE/CVF Conference on Computer Vision and Pattern Recognition}, pages 6405--6414, 2024{\natexlab{b}}.

\bibitem[Dong et~al.(2024{\natexlab{c}})Dong, Zhou, Zhang, Liu, and Chen]{ecmamba24neurips}
Wei Dong, Han Zhou, Yulun Zhang, Xiaohong Liu, and Jun Chen.
\newblock Ecmamba: Consolidating selective state space model with retinex guidance for efficient multiple exposure correction.
\newblock \emph{Advances in Neural Information Processing Systems}, 37:\penalty0 53438--53457, 2024{\natexlab{c}}.

\bibitem[Dong et~al.(2025)Dong, Min, Zhou, and Chen]{sgllie}
Wei Dong, Yan Min, Han Zhou, and Jun Chen.
\newblock Towards scale-aware low-light enhancement via structure-guided transformer design.
\newblock In \emph{Proceedings of the IEEE/CVF Conference on Computer Vision and Pattern Recognition (CVPR) Workshops}, 2025.

\bibitem[Finlayson et~al.(2006)Finlayson, Hordley, Lu, and Drew.]{HirF8}
Graham~D. Finlayson, Steven~D. Hordley, Cheng Lu, and Mark~S. Drew.
\newblock On the removal of shadows from images.
\newblock \emph{IEEE Transactions on Pattern Analysis and Machine Intelligence}, 28\penalty0 (59--68), 2006.

\bibitem[Finlayson et~al.(2009)Finlayson, Drew, and Lu]{HirF9}
Graham~D. Finlayson, Mark~S. Drew, and Cheng Lu.
\newblock Entropy minimization for shadow removal.
\newblock \emph{International Journal of Computer Vision}, 85:\penalty0 35--57, 2009.

\bibitem[Fu et~al.(2021)Fu, Zhou, Guo, Juefei-Xu, Yu, Feng, Liu, and Wang]{fusionnet}
Lan Fu, Changqing Zhou, Qing Guo, Felix Juefei-Xu, Hongkai Yu, Wei Feng, Yang Liu, and Song Wang.
\newblock Auto-exposure fusion for single-image shadow removal.
\newblock In \emph{CVPR}, 2021.

\bibitem[Gong and Cosker(2016)]{HirF14}
Han Gong and Darren~P. Cosker.
\newblock Interactive removal and ground truth for difficult shadow scenes.
\newblock \emph{Journal of the Optical Society of America. A, Optics, image science, and vision}, 33\penalty0 (9):\penalty0 1798--1811, 2016.

\bibitem[Gryka et~al.(2015)Gryka, Terry, and Brostow]{DS14}
Maciej Gryka, Michael Terry, and Gabriel~J Brostow.
\newblock Learning to remove soft shadows.
\newblock \emph{ACM Transactions on Graphics (TOG)}, 34\penalty0 (5):\penalty0 1--15, 2015.

\bibitem[Guo et~al.(2023)Guo, Huang, Liu, Cheng, and Wen]{shadowformer}
Lanqing Guo, Siyu Huang, Ding Liu, Hao Cheng, and Bihan Wen.
\newblock Shadowformer: Global context helps image shadow removal.
\newblock In \emph{AAAI}, 2023.

\bibitem[Guo et~al.(2012)Guo, Dai, and Hoiem]{HirF18}
Ruiqi Guo, Qieyun Dai, and Derek Hoiem.
\newblock Paired regions for shadow detection and removal.
\newblock \emph{IEEE transactions on pattern analysis and machine intelligence}, 35\penalty0 (12):\penalty0 2956--2967, 2012.

\bibitem[He et~al.(2016)He, Zhang, Ren, and Sun]{DRDB14}
Kaiming He, Xiangyu Zhang, Shaoqing Ren, and Jian Sun.
\newblock Deep residual learning for image recognition.
\newblock In \emph{Proceedings of the IEEE conference on computer vision and pattern recognition}, pages 770--778, 2016.

\bibitem[He et~al.(2021)He, Peng, Dong, and Du]{DS18}
Shengfeng He, Bing Peng, Junyu Dong, and Yong Du.
\newblock Maskshadownet: Toward shadow removal via masked adaptive instance normalization.
\newblock \emph{IEEE Signal Processing Letters}, 28\penalty0 (957-961), 2021.

\bibitem[Hu et~al.(2018)Hu, Zhu, Fu, Qin, and Heng]{HirF19}
Xiaowei Hu, Lei Zhu, Chi-Wing Fu, Jing Qin, and Pheng-Ann Heng.
\newblock Direction-aware spatial context features for shadow detection.
\newblock In \emph{Proceedings of the IEEE conference on computer vision and pattern recognition}, pages 7454--7462, 2018.

\bibitem[Huang et~al.(2016)Huang, Wang, Wu, Zhou, and Wu]{tracking}
Guan Huang, Xingang Wang, Wenqi Wu, Han Zhou, and Yuanyuan Wu.
\newblock Real-time lane-vehicle detection and tracking system.
\newblock In \emph{Chinese Control and Decision Conference (CCDC)}, 2016.

\bibitem[Huang et~al.(2017)Huang, Liu, Maaten, and Weinberger]{DRDB15}
Gao Huang, Zhuang Liu, Laurens Van~Der Maaten, and Kilian~Q. Weinberger.
\newblock Densely connected convolutional networks.
\newblock In \emph{Proceedings of the IEEE conference on computer vision and pattern recognition}, pages 4700--4708, 2017.

\bibitem[Jin et~al.(2021)Jin, Sharma, and T.~Tan]{dcshadow}
Yeying Jin, Aashish Sharma, and Robby T.~Tan.
\newblock Dc-shadownet: Single-image hard and soft shadow removal using unsupervised domain-classifier guided network.
\newblock In \emph{ICCV}, 2021.

\bibitem[Land(1977)]{retinex}
Edwin~H Land.
\newblock The retinex theory of color vision.
\newblock \emph{Scientific american}, 237\penalty0 (6):\penalty0 108--129, 1977.

\bibitem[Le and Samaras(2019)]{le2019shadow}
Hieu Le and Dimitris Samaras.
\newblock Shadow removal via shadow image decomposition.
\newblock In \emph{ICCV}, 2019.

\bibitem[Lengyel et~al.(2021)Lengyel, Garg, Milford, and van Gemert]{ciconv}
Attila Lengyel, Sourav Garg, Michael Milford, and Jan~C van Gemert.
\newblock Zero-shot day-night domain adaptation with a physics prior.
\newblock In \emph{Proceedings of the IEEE/CVF International Conference on Computer Vision}, pages 4399--4409, 2021.

\bibitem[Li et~al.(2025)Li, Jin, Jin, Wu, Li, Wang, Yang, Li, Chen, Wen, Tan, Timofte, et~al.]{ntire2025day}
Xin Li, Yeying Jin, Xin Jin, Zongwei Wu, Bingchen Li, Yufei Wang, Wenhan Yang, Yu Li, Zhibo Chen, Bihan Wen, Robby Tan, Radu Timofte, et~al.
\newblock {NTIRE} 2025 challenge on day and night raindrop removal for dual-focused images: Methods and results.
\newblock In \emph{Proceedings of the IEEE/CVF Conference on Computer Vision and Pattern Recognition (CVPR) Workshops}, 2025.

\bibitem[Liang et~al.(2021)Liang, Cao, Sun, Zhang, Gool, and Timofte]{DS22}
Jingyun Liang, Jiezhang Cao, Guolei Sun, Kai Zhang, Luc~Van Gool, and Radu Timofte.
\newblock Swinir: Image restoration using swin transformer.
\newblock In \emph{Proceedings of the IEEE/CVF international conference on computer vision}, pages 1833--1844, 2021.

\bibitem[Lin et~al.(2020)Lin, Chen, and Chuang]{HirF29}
Yun-Hsuan Lin, Wen-Chin Chen, and Yung-Yu Chuang.
\newblock Bedsr-net: A deep shadow removal network from a single document image.
\newblock In \emph{2020 IEEE/CVF Conference on Computer Vision and Pattern Recognition (CVPR)}, pages 12902-- 12911, 2020.

\bibitem[Liu et~al.(2025)Liu, Wu, Vasluianu, Yan, Ren, Zhang, Gu, Zhang, Zhu, Timofte, et~al.]{ntire2025lowlight}
Xiaoning Liu, Zongwei Wu, Florin-Alexandru Vasluianu, Hailong Yan, Bin Ren, Yulun Zhang, Shuhang Gu, Le Zhang, Ce Zhu, Radu Timofte, et~al.
\newblock {NTIRE} 2025 challenge on low light image enhancement: Methods and results.
\newblock In \emph{Proceedings of the IEEE/CVF Conference on Computer Vision and Pattern Recognition (CVPR) Workshops}, 2025.

\bibitem[Liu et~al.(2024)Liu, Ke, Xu, Liu, Wang, and Lau]{DS25}
Yuhao Liu, Zhanghan Ke, Ke Xu, Fang Liu, Zhenwei Wang, and Rynson W.~H. Lau.
\newblock Recasting regional lighting for shadow removal.
\newblock In \emph{Proceedings of the AAAI Conference on Artificial Intelligence}, pages 3810--3818, 2024.

\bibitem[Liu et~al.(2021{\natexlab{a}})Liu, Lin, Cao, Hu, Wei, Zhang, Lin, and Guo]{DS26}
Ze Liu, Yutong Lin, Yue Cao, Han Hu, Yixuan Wei, Zheng Zhang, Stephen Lin, and Baining Guo.
\newblock Swin transformer: Hierarchical vision transformer using shifted windows.
\newblock In \emph{Proceedings of the IEEE/CVF international conference on computer vision}, pages 10012--10022, 2021{\natexlab{a}}.

\bibitem[Liu et~al.(2021{\natexlab{b}})Liu, Yin, Mi, Pu, and Wang]{HirF30}
Zhihao Liu, Hui Yin, Yang Mi, Mengyang Pu, and Song Wang.
\newblock Shadow removal by a lightness-guided network with training on unpaired data.
\newblock \emph{IEEE Transactions on Image Processing}, 30:\penalty0 1853--1865, 2021{\natexlab{b}}.

\bibitem[Liu et~al.(2022)Liu, Mao, Wu, Feichtenhofer, Darrell, and Xie]{convnext}
Zhuang Liu, Hanzi Mao, Chao-Yuan Wu, Christoph Feichtenhofer, Trevor Darrell, and Saining Xie.
\newblock A convnet for the 2020s.
\newblock In \emph{Proceedings of the IEEE/CVF conference on computer vision and pattern recognition}, pages 11976--11986, 2022.

\bibitem[Luo et~al.(2023)Luo, Gustafsson, Zhao, et~al.]{refusion}
Ziwei Luo, Fredrik~K. Gustafsson, Zheng Zhao, et~al.
\newblock Refusion: Enabling large-size realistic image restoration with latent-space diffusion models.
\newblock In \emph{CVPRW}, 2023.

\bibitem[Qu et~al.(2017)Qu, Tian, He, Tang, and Lau]{HirF32}
Liangqiong Qu, Jiandong Tian, Shengfeng He, Yandong Tang, and Rynson W.~H. Lau.
\newblock Deshadownet: A multicontext embedding deep network for shadow removal.
\newblock In \emph{2017 IEEE Conference on Computer Vision and Pattern Recognition (CVPR)}, pages 2308--2316, 2017.

\bibitem[Song et~al.(2023)Song, He, Qian, and Du]{DS35}
Yuda Song, Zhuqing He, Hui Qian, and Xin Du.
\newblock Vision transformers for single image dehazing.
\newblock In \emph{IEEE Transactions on Image Processing}, pages 1927--1941, 2023.

\bibitem[Sun et~al.(2024)Sun, Ren, Gao, Wang, and Cao]{HistoFormer}
Shangquan Sun, Wenqi Ren, Xinwei Gao, Rui Wang, and Xiaochun Cao.
\newblock Restoring images in adverse weather conditions via histogram transformer.
\newblock In \emph{European Conference on Computer Vision}. Cham: Springer Nature Switzerland, 2024.

\bibitem[Vasluianu et~al.(2023)Vasluianu, Seizinger, and Timofte]{vasluianu2023wsrd}
Florin-Alexandru Vasluianu, Tim Seizinger, and Radu Timofte.
\newblock Wsrd: A novel benchmark for high resolution image shadow removal.
\newblock In \emph{CVPRW}, 2023.

\bibitem[Vasluianu et~al.(2024{\natexlab{a}})Vasluianu, Seizinger, Wu, Ranjan, and Timofte]{ifblend24eccv}
Florin-Alexandru Vasluianu, Tim Seizinger, Zongwei Wu, Rakesh Ranjan, and Radu Timofte.
\newblock Towards image ambient lighting normalization.
\newblock In \emph{European Conference on Computer Vision}, pages 385--404. Springer, 2024{\natexlab{a}}.

\bibitem[Vasluianu et~al.(2024{\natexlab{b}})Vasluianu, Seizinger, Zhou, Wu, Chen, Timofte, Dong, Zhou, Tian, Chen, et~al.]{ntire24shadow}
Florin-Alexandru Vasluianu, Tim Seizinger, Zhuyun Zhou, Zongwei Wu, Cailian Chen, Radu Timofte, Wei Dong, Han Zhou, Yuqiong Tian, Jun Chen, et~al.
\newblock Ntire 2024 image shadow removal challenge report.
\newblock In \emph{Proceedings of the IEEE/CVF Conference on Computer Vision and Pattern Recognition}, pages 6547--6570, 2024{\natexlab{b}}.

\bibitem[Vasluianu et~al.(2025)Vasluianu, Seizinger, Zhou, Wu, Chen, Timofte, et~al.]{ntire2025shadow}
Florin-Alexandru Vasluianu, Tim Seizinger, Zhuyun Zhou, Zongwei Wu, Cailian Chen, Radu Timofte, et~al.
\newblock {NTIRE} 2025 image shadow removal challenge report.
\newblock In \emph{Proceedings of the IEEE/CVF Conference on Computer Vision and Pattern Recognition (CVPR) Workshops}, 2025.

\bibitem[Vaswani et~al.(2017)Vaswani, Shazeer, Parmar, Uszkoreit, Jones, Gomez, and Polosukhin]{DS39}
Ashish Vaswani, Noam Shazeer, Niki Parmar, Jakob Uszkoreit, Llion Jones, Aidan~N Gomez, and {\L}ukasz Kaiserand~Illia Polosukhin.
\newblock Attention is all you need.
\newblock In \emph{Advances in Neural Information Processing Systems (NeurIPS)}, 2017.

\bibitem[Wan et~al.(2022)Wan, Yin, Wu, Wu, Liu, and Wang]{HirF39}
J. Wan, Hui Yin, Zhenyao Wu, Xinyi Wu, Y. Liu, and Song Wang.
\newblock Style-guided shadow removal.
\newblock In \emph{European Conference on Computer Vision}, pages 361--378. Cham: Springer Nature Switzerland, 2022.

\bibitem[Wang et~al.(2018)Wang, Li, and Yang]{stcgan}
Jifeng Wang, Xiang Li, and Jian Yang.
\newblock Stacked conditional generative adversarial networks for jointly learning shadow detection and shadow removal.
\newblock In \emph{CVPR}, 2018.

\bibitem[Wang et~al.(2004)Wang, Bovik, Sheikh, and Simoncelli]{DS42}
Zhou Wang, Alan~C. Bovik, Hamid~R. Sheikh, and Eero~P. Simoncelli.
\newblock Image quality assessment: From error visibility to structural similarity.
\newblock \emph{IEEE transactions on image processing}, 13\penalty0 (4):\penalty0 600--612, 2004.

\bibitem[Wang et~al.(2022)Wang, Cun, Bao, Zhou, Liu, and Li]{HistoF75}
Zhendong Wang, Xiaodong Cun, Jianmin Bao, Wengang Zhou, Jianzhuang Liu, and Houqiang Li.
\newblock Uformer: A general u-shaped transformer for image restoration.
\newblock In \emph{Proceedings of the IEEE/CVF conference on computer vision and pattern recognition}, pages 17683--17693, 2022.

\bibitem[Wu et~al.(2020)Wu, Chen, and Tong]{DS43}
Minghu Wu, Rui Chen, and Ying Tong.
\newblock Shadow elimination algorithm using color and texture features.
\newblock \emph{Computational intelligence and neuroscience 2020}, 2020.

\bibitem[Xiao et~al.(2022)Xiao, Fu, Liu, Wu, and Zha]{HistoF78}
Jie Xiao, Xueyang Fu, Aiping Liu, Feng Wu, and Zheng-Jun Zha.
\newblock Image de-raining transformer.
\newblock \emph{IEEE transactions on pattern analysis and machine intelligence}, 45\penalty0 (11):\penalty0 12978--12995, 2022.

\bibitem[Xu et~al.(2022)Xu, Dong, and Zhou]{recognition}
Dong Xu, Wei Dong, and Han Zhou.
\newblock Sclera recognition based on efficient sclera segmentation and significant vessel matching.
\newblock In \emph{The Computer Journal}, 2022.

\bibitem[Xu et~al.(2024)Xu, Lin, Yang, Chao, and Ji]{SADC}
Yimin Xu, Mingbao Lin, Hong Yang, Fei Chao, and Rongrong Ji.
\newblock Shadow-aware dynamic convolution for shadow removal.
\newblock In \emph{Pattern Recognition}, 2024.

\bibitem[Yang et~al.(2025)Yang, Cai, Ouyang, Vasluianu, Timofte, Ding, Sun, Fu, Li, Ho, Meng, et~al.]{ntire2025reflection}
Kangning Yang, Jie Cai, Ling Ouyang, Florin-Alexandru Vasluianu, Radu Timofte, Jiaming Ding, Huiming Sun, Lan Fu, Jinlong Li, Chiu~Man Ho, Zibo Meng, et~al.
\newblock {NTIRE} 2025 challenge on single image reflection removal in the wild: Datasets, methods and results.
\newblock In \emph{Proceedings of the IEEE/CVF Conference on Computer Vision and Pattern Recognition (CVPR) Workshops}, 2025.

\bibitem[Yang et~al.(2012)Yang, Tan, and Ahuja]{HirF48}
Qingxiong Yang, K.~H. Tan, and Narendra Ahuja.
\newblock Shadow removal using bilateral filtering.
\newblock \emph{IEEE Transactions on Image Processing}, 21\penalty0 (10):\penalty0 4361--4368, 2012.

\bibitem[Yu and Koltun(2015)]{DRDB39}
Fisher Yu and Vladlen Koltun.
\newblock Multi-scale context aggregation by dilated convolutions.
\newblock \emph{arXiv preprint arXiv:1511.07122}, 2015.

\bibitem[Yu et~al.(2022)Yu, Dai, Li, Ma, Shen, Li, and Qi]{DRDB}
Xin Yu, Peng Dai, Wenbo Li, Lan Ma, Jiajun Shen, Jia Li, and Xiaojuan Qi.
\newblock Towards efficient and scale-robust ultra-high-definition image demoir´eing.
\newblock In \emph{European Conference on Computer Vision}, pages 646--662. Cham: Springer Nature Switzerland, 2022.

\bibitem[Zamir et~al.(2022)Zamir, Arora, Khan, Hayat, Khan, and Yang]{HistoF86}
Syed~Waqas Zamir, Aditya Arora, Salman Khan, Munawar Hayat, Fahad~Shahbaz Khan, and Ming-Hsuan Yang.
\newblock Restormer: Efficient transformer for high-resolution image restoration.
\newblock In \emph{Proceedings of the IEEE/CVF conference on computer vision and pattern recognition}, pages 5728--5739, 2022.

\bibitem[Zhang et~al.(2018{\natexlab{a}})Zhang, Isola, Efros, Shechtman, and Wang]{DS49}
Richard Zhang, Phillip Isola, Alexei~A Efros, Eli Shechtman, and Oliver Wang.
\newblock The unreasonable effectiveness of deep features as a perceptual metric.
\newblock In \emph{Proceedings of the IEEE Conference on Computer Vision and Pattern Recognition (CVPR)}, 2018{\natexlab{a}}.

\bibitem[Zhang et~al.(2018{\natexlab{b}})Zhang, Tian, Kong, Zhong, and Fu]{DRDB46}
Yulun Zhang, Yapeng Tian, Yu Kong, Bineng Zhong, and Yun Fu.
\newblock Residual dense network for image super-resolution.
\newblock In \emph{Proceedings of the IEEE conference on computer vision and pattern recognition}, pages 2472--2481, 2018{\natexlab{b}}.

\bibitem[Zhao et~al.(2023)Zhao, Gou, Li, Peng, Lv, and Peng]{HistoF96}
Haiyu Zhao, Yuanbiao Gou, Boyun Li, Dezhong Peng, Jiancheng Lv, and Xi Peng.
\newblock Comprehensive and delicate: An efficient transformer for image restoration.
\newblock In \emph{Proceedings of the IEEE/CVF conference on computer vision and pattern recognition}, pages 14122--14132, 2023.

\bibitem[Zhao et~al.(2018)Zhao, Elliott, Zhou, and Rafferty]{DS50}
Yunfeng Zhao, Chris Elliott, Huiyu Zhou, and Karen Rafferty.
\newblock Pixel-wise illumination correction algorithms for relative color constancy under the spectral domain.
\newblock In \emph{IEEE International Symposium on Signal Processing and Information Technology (ISSPIT)}, 2018.

\bibitem[Zhou et~al.(2023)Zhou, Dong, Liu, and Chen]{zhou2023breaking}
Han Zhou, Wei Dong, Yangyi Liu, and Jun Chen.
\newblock Breaking through the haze: An advanced non-homogeneous dehazing method based on fast fourier convolution and convnext.
\newblock In \emph{Proceedings of the IEEE/CVF Conference on Computer Vision and Pattern Recognition}, pages 1895--1904, 2023.

\bibitem[Zhou et~al.(2024)Zhou, Dong, Liu, Liu, Min, Zhai, and Chen]{glare24eccv}
Han Zhou, Wei Dong, Xiaohong Liu, Shuaicheng Liu, Xiongkuo Min, Guangtao Zhai, and Jun Chen.
\newblock Glare: Low light image enhancement via generative latent feature based codebook retrieval.
\newblock In \emph{European Conference on Computer Vision}, pages 36--54. Springer, 2024.

\bibitem[Zhou et~al.(2025{\natexlab{a}})Zhou, Dong, and Chen]{lita25cvpr}
Han Zhou, Wei Dong, and Jun Chen.
\newblock Lita-gs: Illumination-agnostic novel view synthesis via reference-free 3d gaussian splatting and physical priors.
\newblock \emph{arXiv preprint arXiv:2504.00219}, 2025{\natexlab{a}}.

\bibitem[Zhou et~al.(2025{\natexlab{b}})Zhou, Dong, Liu, Zhang, Zhai, and Chen]{gpp25aaai}
Han Zhou, Wei Dong, Xiaohong Liu, Yulun Zhang, Guangtao Zhai, and Jun Chen.
\newblock Low-light image enhancement via generative perceptual priors.
\newblock In \emph{Proceedings of the AAAI Conference on Artificial Intelligence}, pages 10752--10760, 2025{\natexlab{b}}.

\bibitem[Zhu et~al.(2022)Zhu, Xiao, Fang, Fu, Xiong, and Zha]{bmnet}
Yurui Zhu, Zeyu Xiao, Yanchi Fang, Xueyang Fu, Zhiwei Xiong, and Zheng-Jun Zha.
\newblock Efficient model-driven network for shadow removal.
\newblock In \emph{AAAI}, 2022.

\end{thebibliography}
}

\end{document}